\documentclass[runningheads]{llncs}
\usepackage{float}
\usepackage{graphicx}
\usepackage{multirow}
\usepackage{booktabs}
\usepackage{amsmath}
\usepackage{bbm}
\usepackage[colorlinks, citecolor=blue, anchorcolor=blue, linkcolor=blue]{hyperref}
\makeatletter
\newif\if@restonecol
\makeatother

\usepackage[linesnumbered,ruled,vlined]{algorithm2e}
\usepackage{algpseudocode}
\usepackage{amsmath}
\usepackage[misc]{ifsym}

\begin{document}
%
\title{Improving Knowledge Graph Entity Alignment with Graph Augmentation}

\titlerunning{GAEA}
%
\author{Feng Xie, Xiang Zeng, Bin Zhou$^{(\textrm{\Letter})}$, and Yusong Tan}
\authorrunning{F. Xie et al.}
%
\institute{College of Computer, National University of Defense Technology, Changsha, China
\email{\{xiefeng,zengxiang,binzhou,ystan\}@nudt.edu.cn}}
\maketitle              
\begin{abstract}
Entity alignment (EA) which links equivalent entities across different knowledge graphs (KGs) plays a crucial role in knowledge fusion. In recent years, graph neural networks (GNNs) have been successfully applied in many embedding-based EA methods. However, existing GNN-based methods either suffer from the structural heterogeneity issue that especially appears in the real KG distributions or ignore the heterogeneous representation learning for unseen (unlabeled) entities, which would lead the model to overfit on few alignment seeds (i.e., training data) and thus cause unsatisfactory alignment performance. To enhance the EA ability, we propose GAEA, a novel EA approach based on graph augmentation. In this model, we design a simple Entity-Relation (ER) Encoder to generate latent representations for entities via jointly modeling comprehensive structural information and rich relation semantics. Moreover, we use graph augmentation to create two graph views for margin-based alignment learning and contrastive entity representation learning, thus mitigating structural heterogeneity and further improving the model's alignment performance. Extensive experiments conducted on benchmark datasets demonstrate the effectiveness of our method. Our codes are available at \url{https://github.com/Xiefeng69/GAEA}. 

\end{abstract}
\section{Introduction}
Knowledge graphs (KGs) can effectively organize and represent facts about the world in a structured fashion.  More and more KGs have been constructed based on different data sources or for different purposes. Therefore, the knowledge contained in different KGs is far from complete yet complementary \cite{xin2022informed}. Entity alignment (EA) which aims to link semantically equivalent entities located on different KGs has attracted increasing attention since it could facilitate knowledge integration and thus promote knowledge-driven applications, such as question answering, recommender systems, and semantic search.

In recent years, embedding-based EA methods \cite{chen2016mtranse,zhu2017iptranse,wang2018gcnalign,guo2019rsn,sun2020alinet,sun2020hyperka,yu2021kegcn,xin2022informed} have achieved decent results. The general pipeline can be summarized into two steps: (\uppercase\expandafter{\romannumeral1}) generating low-dimensional embeddings (latent representations) for entities via KG encoder (e.g., TransE \cite{bordes2013transe}), and then (\uppercase\expandafter{\romannumeral2}) pulling two KGs into a unified embedding space through prior alignment seeds and pairing each entity by distance metrics (e.g., Euclidean distance). Moreover, some works further improve the EA performance by introducing extra information, such as entity names \cite{zhang2019multike}, attributes \cite{liu2020attrgnn,sun2022revisiting}, and literal descriptions \cite{yang2019hman}, while these discriminative features are usually privacy sensitive, noise polluted, and hard to collect \cite{pei2022noise}.

Due to the powerful structure learning capability, Graph Neural Networks (GNNs) like GCN \cite{kipf2016gcn} and GAT \cite{velivckovic2017gat} have been employed as the encoder with Siamese architecture (i.e., shared-parameter) for many embedding-based models \cite{wang2018gcnalign,sun2020alinet,mao2020mraea,yu2021kegcn}. KGs are heterogeneous, especially in real KG distributions, which means entities that have the same referent in different KGs usually have dissimilar relational neighborhood. To address this problem, existing GNN-based models modify and improve GNN variants to better capture structural information in KGs, e.g., AliNet \cite{sun2020alinet} adopts multi-hop aggregation with gating mechanism to expand neighborhood ranges and RDGCN \cite{wu2019rdgcn} incorporates relation features via attention interactions for embedding learning. However, these models introduce a large number of neural network operations and ignore representation learning for unseen entities, which will tend to make the models overfit on few alignment seeds and thus undermine their generalization and performance.

In this paper, we propose GAEA, a novel knowledge graph entity alignment model based on graph augmentation. Firstly, we design an Entity-Relation (ER) Encoder to generate entity representations via jointly leveraging neighborhood structures and relation semantics in KGs. Then, we apply graph augmentation to increase the structural diversity of input KG in the alignment learning process, which encourages the model to capture the semantic importance of different neighbors and enforces the model to obtain stable representations against structure perturbation, thus mitigating overfitting issue to some extent. Moreover, since graph augmentation can inherently generate two distinct graph views without extra parameters, we can let the model perceive structural differences and further improve the feature learning for (unseen) entities by applying contrastive entity representation learning to maximize the consistency between the original KG and augmented KG \cite{you2020gcl,wan2021contrastive}. Our experiments on benchmark datasets OpenEA \cite{sun2020benchmarking} show that GAEA outperforms the existing state-of-the-art embedding-based EA methods. We also conduct thorough auxiliary analyses to demonstrate the effectiveness of incorporating graph augmentation techniques.

\section{Related Works}
Entity alignment is a fundamental task to identify the same entities across different KGs, which has attracted increasing attention in recent years. The existing embedding-based methods can be roughly divided into two categories:

\begin{enumerate}
    \item \textbf{Structure-based models.} These models solely rely on the original structure information of KGs (i.e., triples) to align entities. Previous methods mainly use knowledge representation learning to generate low-dimensional embeddings for entities \cite{chen2016mtranse,zhu2017iptranse,pei2019sea}. For example, MTransE \cite{chen2016mtranse} applies TransE \cite{bordes2013transe} to embed different KGs into independent vector spaces and constructs transitions via proposed alignment modules. Inspired by the powerful structure learning ability of Graph Neural Networks (GNNs), a large body of works begin to focus on employing GNNs as the encoder. GCN-Align \cite{wang2018gcnalign} incorporates GCN \cite{kipf2016gcn} to capture entities' neighborhood structures for the first time and achieves promising results. Subsequent works not only apply various GNN variants, like GAT \cite{velivckovic2017gat}, but also improve the structure awareness by overcoming heterogeneity of different KGs \cite{sun2020alinet,guo2019rsn,wu2019rdgcn}, capturing multi-context structural features \cite{xin2022informed}, and infusing relation semantics \cite{mao2020mraea,sun2020hyperka}.
    \item \textbf{Enhancement-based models.} These models aim to build a high-accuracy alignment system using designed alignment strategies or extra information. BootEA \cite{sun2018bootea} applies iterative learning to find potential alignments and adds them to the training set for data augmentation. CEA \cite{zeng2020cea} formulates alignment inference as a stable matching problem to model collective signals, successfully guaranteeing 1-to-1 alignment. Other effective models introduce extra information to enhance the alignment performance, including entity names \cite{zhang2019multike}, attributes \cite{liu2020attrgnn,sun2022revisiting}, and literal descriptions \cite{yang2019hman}.
\end{enumerate}

In this work, we aim to improve the performance and efficiency of entity alignment only utilizing structural contexts which are abundant and always available without privacy issues in the real-world KGs.

\section{Preliminaries}
\paragraph{\rm \textbf{Knowledge graph.}} A knowledge graph (KG) is formalized as $G=(E,R,T)$, where $E$ and $R$ refer to the set of entities and the set of relations, respectively. $T=E\times{R}\times{E}=\{(h,r,t)|h,t\in{E}\wedge r\in{R}\}$ is the set of triples, where $h$, $r$, and $t$ denote the head entity, connected relation, tail entity, respectively.

\paragraph{\rm \textbf{Entity alignment.}} Given two KGs: $G_s=(E_s,R_s,T_s)$ as the source KG and $G_t=(E_t,R_t,T_t)$ as the target KG, and few alignment seeds  (aka pre-aligned entity pairs) $S=\{(e_i,e_j)|e_i\in{E_s}\wedge e_j\in{E_t} \wedge e_i\equiv{e_j}\}$, where $\equiv$ means equivalence relationship, entity alignment (EA) aims to seek remaining equivalent entities located on different KGs via entity representations.

\paragraph{\rm \textbf{Augmented graph.}} Graph augmentation techniques will generate a perturbed version of the original graph, i.e., augmented graph, by augmentation strategies (e.g., node dropping, edge perturbation). In order not to introduce wrong facts, we only choose edge dropping in this work. At each training iteration, we randomly drop out some triples based on the deletion ratio $r\sim uniform(0,pr)$, where $pr$ is a preset upper bound of the deletion ratio. The augmented graphs for $G_s$ and $G_t$ are denoted as $G_s^{aug}$ and $G_t^{aug}$, respectively. Note that we do not consider deleting the triples associated with entities whose degree is less than 2, because these long-tail entities have sparse neighborhood structures inherently.

\begin{figure}
    \centering
    \includegraphics[width=0.98\linewidth]{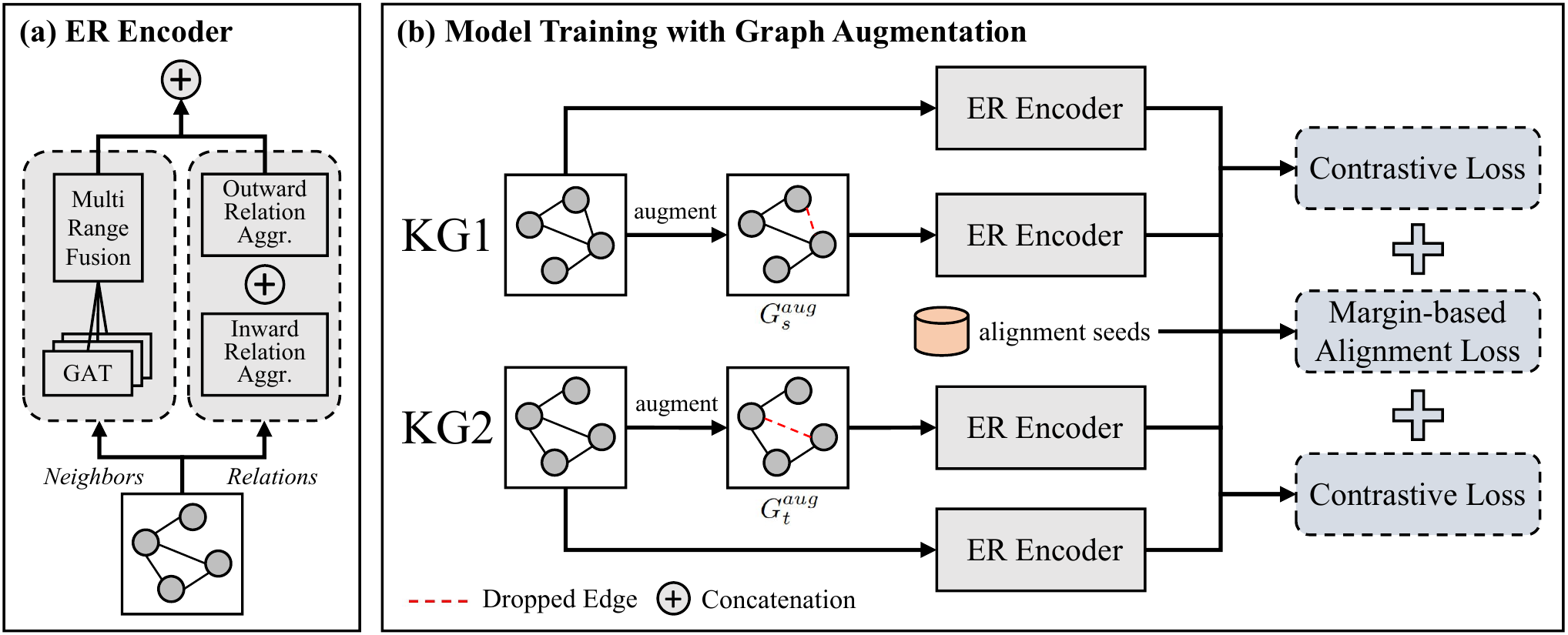}
    \caption{The framework of our proposed GAEA.}
    \label{fig:framework}
\end{figure}

\section{Methodology}
This section details our proposed method, termed as GAEA, which is drawn in Fig. \ref{fig:framework}: (a) Entity-Relation (ER) Encoder which generates latent representations for entities by capturing neighborhood structures and relation semantics jointly; (b) the training process of GAEA can be decomposed into multiple epochs, and in each epoch, we incorporate graph augmentation to conduct margin-based alignment learning and contrastive entity representation learning.

\textbf{Initialization.} At the beginning, we randomly initialize entity embeddings $\textbf{H}^{ent}\in\mathbbm{R}^{(|E_s|+|E_t|)\times{d_{ent}}}$ and relation embeddings $\textbf{H}^{rel}\in\mathbbm{R}^{|R_s\cup{R_t}|\times{d_{rel}}}$, where $d_{ent}$ and $d_{rel}$ are the embedding dimension of entities and relations, respectively. 

\subsection{Entity-Relation Encoder}
Here, we present the Entity-Relation Encoder (ER Encoder for short), which aims to fully capture the contextual information of entities using two aspects jointly: (\uppercase\expandafter{\romannumeral1}) neighborhood structures and (\uppercase\expandafter{\romannumeral2}) relation semantics.

\textbf{Neighborhood aggregator.} First, we aggregate neighbor entities’ information to the central entity. The rationality of neighborhood aggregator lies in the structure assumption that, equivalent entities tend to have similar neighbor structures \cite{wang2018gcnalign}. Moreover, leveraging multi-range neighborhood structures is capable of providing more alignment evidence and mitigating the structural heterogeneity issue. In this work, we apply Graph Attention Network (GAT) \cite{velivckovic2017gat} to allow the central entity to learn the importance of different neighbors and thus selectively aggregate surrounding information, and we then recursively capture multi-range neighbor information by stacking multiple layers:
\begin{equation}
    \textbf{h}_{e_i}^{(l)}=\sum_{e_j\in{N_{e_i}}}\alpha_{ij} \textbf{h}_{e_j}^{(l-1)},
\end{equation}
\begin{equation}
    \alpha_{ij} = \frac{\text{exp}(\text{LeakyReLU}(\textbf{a}^{\top}[\textbf{W}_g\textbf{h}_{e_i}\oplus\textbf{W}_g\textbf{h}_{e_j}]))}{\sum_{{e_k}\in{N_{e_i}}}\text{exp}(\text{LeakyReLU}(\textbf{a}^{\top}[\textbf{W}_g\textbf{h}_{e_i}\oplus\textbf{W}_g\textbf{h}_{e_k}]))},
\end{equation}
where $\top$ represents transposition, $\oplus$ is the concatenation operation, $\textbf{W}_g$ and $\textbf{a}$ are the transformation parameter and attention transformation vector, respectively. $N_{e_i}$ means the neighbor set of entity $e_i$ in KG, and $\alpha_{ij}$ indicates the learned importance of entity $e_j$ to entity $e_i$. $\textbf{h}_{e_i}^{(l)}$ denotes the embedding of $e_i$ at $l$-th layer (total $L$ layers) with $\textbf{H}^{(0)}=\textbf{H}^{ent}$. Note that here we remove the feature transformation and nonlinear activation that act on input embeddings in vanilla GAT since we mainly focus on information aggregation. We only use $\textbf{W}_g$ and $\textbf{a}$ to make each entity aware of its neighborhood contexts.

After multi-layer GAT, we obtain the multi-range neighborhood structural representation matrix for each entity, i.e., $\textbf{H}_{e_i}^m=[\textbf{h}_{e_i}^{(1)},...,\textbf{h}_{e_i}^{(L)}]\in\mathbbm{R}^{L\times{d_{ent}}}$ for $e_i$. Since different neighborhood ranges have different contributions to characterize the central entity, it is necessary to employ a mechanism to adaptively control the flow of each range and thus reduce noise. Inspired by the skipping connections in neural networks \cite{srivastava2015highway,xu2018jk,sun2020alinet}, we firstly utilize a Scaled Dot-Product Attention mechanism \cite{vaswani2017attention} to learn the importance of each range, and then fuse small-range and wide-range representations by weighted average:
\begin{equation}
    [\hat{\textbf{h}}_{e_i}^{(1)},...,\hat{\textbf{h}}_{e_i}^{(L)}] = \text{softmax}(\frac{(\textbf{H}^m_{e_i}\textbf{W}_q)(\textbf{H}^m_{e_i}\textbf{W}_k)^{\top}}{\sqrt{d_{ent}}})\textbf{H}^m_{e_i}
\end{equation}
\begin{equation}
    \textbf{h}_{e_i}^{n} = \frac{1}{L}\sum_{l=1}^{L}\hat{\textbf{h}}_{e_i}^{(l)},
\end{equation}
where $1/\sqrt{d_{ent}}$ is the scaling factor, $\textbf{W}_q$ and $\textbf{W}_k$ are the learnable parameter matrices, and $\textbf{h}_{e_i}^{n}$ is the output of neighborhood aggregator.

\textbf{Relation aggregator.} Relation-level information which carries rich semantics is vital to align entities in KGs \cite{zhang2019multike,yang2019hman} because two equivalent entities may share overlapping relations. MRAEA \cite{mao2020mraea} pointed out that relation directions impose extra but delicate constraints on the head and tail entity individually. Therefore, in this work, we directly use two mean aggregators to gather outward relation semantics and inward relation semantics separately to provide supplementary alignment signals for heterogeneous KGs:
\begin{equation}
    \textbf{h}_{e_i}^r=\frac{1}{|N_{e_i}^{r+}|}\sum_{r\in{N_{e_i}^{r+}}}\textbf{h}_{r}^{rel} \oplus \frac{1}{|N_{e_i}^{r-}|}\sum_{r\in{N_{e_i}^{r-}}}\textbf{h}_{r}^{rel},
\end{equation}
where $N_{e_i}^{r+}$ and $N_{e_i}^{r-}$ are the outward and inward relation set of $e_i$, respectively. 

\textbf{Feature fusion.} Finally, we concatenate two aspects of information:
\begin{equation}
    \tilde{\textbf{h}}_{e_i} =  \textbf{h}_{e_i}^n \oplus \textbf{h}_{e_i}^r,
\end{equation}
where $\tilde{\textbf{h}}_{e_i}\in\mathbbm{R}^{d_{ent}+2\times{d_{rel}}}$ is the final output representation of ER Encoder for $e_i$. In the following training process, the ER Encoder is shared for $G_s$, $G_t$, and their augmented graphs, and given an entity $e_i$, we denote by $\tilde{\textbf{h}}_{e_i}$ its representation generated by ER Encoder with the original graph as input, and $\tilde{\textbf{h}}_{e_i}^{aug}$ its representation generated with the augmented graph as input.

\subsection{Model Training with Graph Augmentation}
Graph augmentation learning has been demonstrated to promote the performance of graph learning, such as overcoming overfitting and oversmoothing issues \cite{rong2019dropedge}, and being used for graph contrastive learning \cite{you2020gcl}. We apply graph augmentation for EA and highlight two main enhancements contributed by it: (\uppercase\expandafter{\romannumeral1}) injecting perturbations into the original KG can increase the diversity of the structural differences, thus preventing the model from overfitting to the training data during alignment process to some extent as well as enforcing the model to produce robust entity representations against structural changes; (\uppercase\expandafter{\romannumeral2}) graph augmentation inherently generates two graph views without extra parameters, which facilitates conducting contrastive learning to promote heterogeneous representation learning for (unseen) entities by contrasting different views.

\textbf{Margin-based alignment loss.} In order to make equivalent entities close to each other and unmatched entities pull away from each other in a unified embedding space. Following previous works \cite{wang2018gcnalign,mao2020mraea,liu2020attrgnn}, we apply the margin-based alignment loss supervised by pre-aligned entity pairs $S$. Notably, here, we use the output of ER Encoder based on augmented graphs to make the model avoid overfitting and behave durable against edge changes:
\begin{equation}\label{eq:align-loss}
    \mathcal{L}_{a}=\sum_{(e_i,e_j)\in{S}} \sum_{(\bar{e}_{i},\bar{e}_{j})\in{\bar{S}_{(e_i,e_j)}}} \left[||\tilde{\textbf{h}}_{e_i}^{aug}-\tilde{\textbf{h}}_{e_j}^{aug}||_{L2}+\rho-||\tilde{\textbf{h}}_{\bar{e}_i}^{aug}-\tilde{\textbf{h}}_{\bar{e}_j}^{aug}||_{L2}\right]_{+},
\end{equation}
where $\rho$ is a hyper-parameter of margin, $[x]_{+}=\text{max}\{0,x\}$ is to ensure non-negative output, and $\bar{S}_{(e_i,e_j)}$ denotes the set of negative entity alignments constructed by corrupting the ground-truth alignment $(e_i,e_j)$, i.e., replacing $e_i$ or $e_j$ with another entity in $G_s$ or $G_t$ via negative sampling strategy.

\textbf{Contrastive loss.} 
Contrastive learning is a good means to explore supervision signals from the vast unlabeled data. Many graph learning works \cite{velivckovic2018DGI,you2020gcl,wan2021contrastive} apply it to learn representations by contrasting different views and then maximizing feature consistency between them. RAC \cite{zeng2021rac} is an effective EA model which incorporates contrastive learning to ameliorate the alignment performance. However, RAC needs to employ two separate graph encoders with the same architecture to model different views of the structural features of entities, which will bring twice the parameters and damage the diversity of graph views. Graph augmentation inherently provides two different views (i.e., original graph view and augmented graph view) without extra parameters. Therefore, we define the contrastive loss to improve entity representation learning by maximizing the feature consistency between the original structure and augmented structure:
\begin{equation}\label{eq:contrastive-loss1}
    \mathcal{L}_{c}=\sum_{z=\{s,t\}}\frac{1}{2|E_z|}\sum_{e_i\in{E_z}}(\mathcal{L}_{c,e_i}^{(G_z,G^{aug}_z)} + \mathcal{L}_{c,e_i}^{(G^{aug}_z,G_z)}),
\end{equation}
\begin{equation}\label{eq:contrastive-loss2}
    \mathcal{L}_{c,e_i}^{(G_z,G_z^{aug})}=-\text{log}\frac{\text{exp}(\langle\text{proj}(\tilde{\textbf{h}}_{e_i}),\text{proj}(\tilde{\textbf{h}}_{e_i}^{aug})\rangle)}{\sum_{e_k\in{E_z}}\text{exp}(\langle \text{proj}(\tilde{\textbf{h}}_{e_i}),\text{proj}(\tilde{\textbf{h}}_{e_k}^{aug})\rangle)},
\end{equation}
where $\langle\cdot\rangle$ means inner product, and $\text{proj}(\cdot)$ is a shared projection head consisting of a linear layer and a ReLU activation function to map entity representations to low-dimensional vector space \cite{you2020gcl}. The definition of the symmetric contrastive loss term $\mathcal{L}_{c,e_i}^{(G^{aug}_z,G_z)}$ is similar with Eq.(\ref{eq:contrastive-loss2}).

\textbf{Model training.} We combine the margin-based alignment loss and the contrastive loss, arriving at the final objective of our model:
\begin{equation}
    \mathcal{L} = \mathcal{L}_a + \lambda\mathcal{L}_c,
\end{equation}
where $\lambda\ge0$ is a tunable parameter weighting the two objectives. The training process of GAEA is outlined in Algorithm \ref{ref:algorithm}, where negative sample set and augmented graphs will be updated every iteration (10 epochs as an iteration).

\begin{algorithm}\label{ref:algorithm}
    \caption{Training Procedure of GAEA}
    \LinesNumbered
    \KwIn{Knowledge graph $G_s$ and $G_t$, pre-aligned entity pairs $S$.}
    Initialize entity embeddings and relation embeddings;\\
    \While{Not Converge}
    {
        \For{each Epoch}
        {
            \If(\tcp*[f]{10 epochs as an iteration}){Epoch \% 10 == 0}
            {
                Generate augmented graphs $G_s^{aug}$ and $G_t^{aug}$ for $G_s$ and $G_t$;\\
                Generate negative sample set $\bar{S}$ based on $S$;\\
            }
            Generate entity representations using ER Encoder; \\
            Calculate $\mathcal{L}_a$ using $S$ and $\bar{S}$ via Eq.(\ref{eq:align-loss});\\
            Calculate $\mathcal{L}_c$ using Eq.(\ref{eq:contrastive-loss1}) and Eq.(\ref{eq:contrastive-loss2});\\
            $\Theta\gets\text{BackProp}(\mathcal{L}_a+\lambda\mathcal{L}_c)$; \Comment{Adam step}\\
        }
    }
    \textbf{return} Model parameters $\Theta$;
\end{algorithm}
\vspace{-6pt}

\subsection{Alignment Inference}
After pulling embeddings from two KGs into a unified vector space and making them comparable, alignment relationships can be inferred by measuring the distance between two entities. In this work, we use Euclidean Distance to be the distance metric, i.e., for $e_i\in{E_s}$ and $e_j\in{E_t}$, the distance between entity pair ($e_i$,$e_j$) is calculated by $|| \tilde{\textbf{h}}_{e_i} - \tilde{\textbf{h}}_{e_j} ||_{L2}$. In order to find $e_i$' alignment relationship, we calculate its distance to all entities belonging to $G_t$ and perform the nearest neighbor (NN) search to identify $e_i$' counterpart entity in $G_t$:
\begin{equation}
    e_j=\text{arg}\mathop{min}\limits_{e_j^{'}\in{E_t}}|| \tilde{\textbf{h}}_{e_i} - \tilde{\textbf{h}}_{e_j^{'}} ||_{L2}.
\end{equation}

Notably, we use the original KG structures in the inference phase instead of augmented versions to generate final entity representation $\tilde{\textbf{h}}$. We apply Faiss\footnote{https://github.com/facebookresearch/faiss} to accelerate the alignment inference process.

\section{Experimental Setup}

\subsection{Experimental Setup}

\paragraph{\rm \textbf{Datasets.}}We use the 15K benchmark dataset (V1) in OpenEA \cite{sun2020benchmarking} for evaluation since the entities thereof follow the degree distribution in real-world KGs. It contains two cross-lingual settings, i.e., EN-FR-15K (English-to-French) and EN-DE-15K (English-to-German), and two monolingual settings, i.e., D-W-15K (DBPedia-to-Wikidata) and D-Y-15K (DBPedia-to-YAGO). Following the data splits in OpenEA, we use the same split setting where 20\%, 10\%, and 70\% alignments are harnessed for training, validation, and testing, respectively.

\paragraph{\rm \textbf{Metrics.}}We adopt \textit{Hits@k} ($k$=1,5) and \textit{Mean Reciprocal Rank} (\textit{MRR}) as the evaluation metrics. \textit{Hits@k} is to measure the alignment accuracy, while \textit{MRR} measures the average performance of ranking over all test samples. The higher the \textit{Hits@k} and \textit{MRR}, the better the alignment performance.

\paragraph{\rm \textbf{Baselines.}}We choose some GNN variants and several existing state-of-the-art embedding-based EA models as baselines: GCN \cite{kipf2016gcn} and GAT \cite{velivckovic2017gat} are the classic variants of GNNs; MTransE \cite{chen2016mtranse} and SEA \cite{pei2019sea} are triple-based methods that capture the local semantics information of relation triples via knowledge representation learning; GCN-Align \cite{wang2018gcnalign}, AliNet \cite{sun2020alinet}, HyperKA \cite{sun2020hyperka}, and KE-GCN \cite{yu2021kegcn} are the neighborhood-based methods which apply GNNs to explore neighborhood structure information; IPTransE \cite{zhu2017iptranse} and RSNs \cite{guo2019rsn} both are path-based methods that extract the long-term dependencies across relation paths; IMEA \cite{xin2022informed} is the recent strong baseline which uses Transformer-like architecture to capture multiple structural contexts in an end-to-end manner.

We should note here that our model and the above baselines all mainly focus on the structural information of KGs. Therefore, for a fair comparison, we do not consider the models which utilize extra information (e.g., attributes, literals) for enhancement, such as AttrGNN \cite{liu2020attrgnn}, HMAN \cite{yang2019hman}, MultiKE \cite{zhang2019multike}.

\begin{table*}[htbp]
  \renewcommand{\arraystretch}{1.3} 
  \centering
  \scriptsize
  \caption{Entity alignment results in cross-lingual and monolingual settings. The results with $\dagger$ are retrieved from \cite{sun2020benchmarking}, and $\ddagger$ from \cite{xin2022informed}. Results labeled by $\ast$ are reproduced using the released source codes. The \textbf{boldface} indicates the best result of each column and \underline{underlined} the second-best.}
  \setlength{\belowcaptionskip}{0pt}
  \label{tab:performance-comparison}
  \setlength{\tabcolsep}{0.7mm}{
  \begin{tabular}{lcccccccccccc}
    \hline
    \multirow{2.5}{*}{Models} & \multicolumn{3}{c}{EN-FR-15K} & \multicolumn{3}{c}{EN-DE-15K} & \multicolumn{3}{c}{D-W-15K} & \multicolumn{3}{c}{D-Y-15K}\\
    \cmidrule(r){2-4} \cmidrule(r){5-7} \cmidrule(r){8-10} \cmidrule(r){11-13}
    & Hit@1 & Hit@5 & MRR & Hit@1 & Hit@5 & MRR & Hit@1 & Hit@5 & MRR & Hit@1 & Hit@5 & MRR\\
    \hline
    GCN$^\ast$ & .210 & .414 & .304 & .304 & .497 & .394 & .208 & .367 & .284 & .343 & .503 & .416\\
    GAT$^\ast$ & .297 & .585 & .426 & .542 & .737 & .630 & .383 & .622 & .489 & .468 & .707 & .573\\
    MTrasnE$^\dagger$ & .247 & .467 & .351 & .307 & .518 & .407 & .259 & .461 & .354 & .463 & .675 & .559\\
    SEA$^\dagger$ & .280 & .530 & .397 & .530 & .718 & .617 & .360 & .572 & .458 & .500 & .706 & .591\\
    IPTransE$^\dagger$ & .169 & .320 & .243 & .350 & .515 & .430 & .232 & .380 & .303 & .313 & .456 & .378\\
    RSNs$^\dagger$ & .393 & .595 & .487 & .587 & .752 & .662 & .441 & .615 & .521 & .514 & .655 & .580\\
    GCN-Align$^\dagger$ & .338 & .589 & .451 & .481 & .679 & .571 & .364 & .580 & .461 & .465 & .626 & .536\\
    AliNet$^\ddagger$ & .364 & .597 & .467 & .604 & .759 & .673 & .440 & .628 & .522 & .559 & .690 & .617\\
    HyperKA$^\ddagger$ & .353 & .630 & .477 & .560 & .780 & .656 & .440 & .686 & .548 & .568 & .777 & .659\\
    KE-GCN$^\ddagger$ & .408 & .670 & .524 & \underline{.658} & .822 & \underline{.730} & .519 & .727 & .608 & .560 & .750 & .644\\
    IMEA$^\ddagger$ & \underline{.458} & \underline{.720} & \underline{.574} & .639 & \underline{.827} & .724 & \underline{.527} & \underline{.753} & \underline{.626} & \textbf{.639} & \textbf{.804} & \textbf{.712}\\
    \hline
    GAEA & \textbf{.486} & \textbf{.746} & \textbf{.602} & \textbf{.684} & \textbf{.854} & \textbf{.760} & \textbf{.562} & \textbf{.768} & .\textbf{654} & \underline{.608} & \underline{.791} & \underline{.688}\\
    w/o $rel.$ & .324 & .626 & .458 & .593 & .785 & .678 & .409 & .666 & .521 & .502 & .743 & .605\\
    \hline
  \end{tabular}}
\end{table*}

\paragraph{\rm \textbf{Implementation details.}}All programs are implemented using \textit{Python} 3.6.13 and \textit{PyTorch} 1.10.2 with \textit{CUDA} 11.3 on an \textit{NVIDIA GeForce RTX 3090} GPU. Following OpenEA \cite{sun2020benchmarking}, we report the average results of five-fold cross-validation. We initialize trainable parameters with the Xavier initializer, and we train the model using Adam optimizer with weight decay 1e-5 and perform early stopping to terminate training based on the \textit{MRR} score tested every 10 epochs on the validation data. As for hyper-parameters, the learning rate is set to 0.001, the dropout rate is 0.2, the layer number of GAT $L$ is 2, the number of negative samples for each entity is 5, the negative sampling strategy is $\epsilon$-Truncated Uniform Negative Sampling \cite{sun2018bootea} with $\epsilon$ = 0.9, the margin $\rho$ is 1, the balance parameter $\lambda$ is 100, and the embedding dimension of entities $d_{ent}$ and relations $d_{rel}$ are set to 256 and 128, respectively. The $pr$ is searched in \{0.05, 0.1, 0.15\}. Following the convention, the default alignment direction is from left to right. Taking D-W-15K as an example, we regard DBpedia as the source KG and seek to find the counterparts of source entities in the target KG Wikidata.

\subsection{Experimental Results}
\paragraph{\rm \textbf{Performance comparison.}} Table \ref{tab:performance-comparison} reports the comparison results on the OpenEA 15K datasets. Experimental results show that our proposed GAEA outperforms other models in most tasks, especially in cross-lingual settings. There is a phenomenon that the performance of models utilizing knowledge representation learning as the encoder, e.g., MTransE, SEA, and IPTransE, are inferior compared with the models applying GNNs as the encoder like AliNet and KE-GCN, and have on-par or even worse performance than vanilla GCN and GAT, which demonstrates the GNNs' powerful representation ability in EA. We also notice that, compared with some methods applying GCN as the encoder (e.g., GCN-Align, AliNet), the vanilla GCN fails to surpass them, which shows the significance of designing a more effective encoder for representing entities in KGs. IMEA is a strong baseline that captures abundant structure contexts and it obtains excellent results on D-Y-15K task. However, IMEA introduces carefully designed data processing (e.g., entity paths encoding) and becomes a complicated network due to the Transformer-like architecture, which will inevitably increase the training difficulty and overfitting risk. Additionally, we compare the model size (denoted as \#Params) in Table \ref{tab:params-comparison}. GAEA greatly reduces the number of parameters compared to IMEA while acquiring decent alignment performance. This is because GAEA designs a simple Entity-Relation Encoder to capture multi-range neighborhood structures to mitigate heterogeneity and infuse relation semantics to provide more comprehensive signals for alignment. Moreover, GAEA further facilitates producing expressive and robust entity representations by integrating graph augmentation to achieve alignment learning supervised by alignment seeds and contrastive representation learning for unseen entities. In summary, our proposed GAEA is a light and powerful solution for EA.

\begin{minipage}{0.95\textwidth}
    \centering
    \begin{minipage}[t]{0.4\textwidth}
        \begin{table}[H]
            \centering
            \scriptsize
            \caption{\#Params comparison.}
            \label{tab:params-comparison}
            \begin{tabular}{lr}
                \midrule
                Models     &  \#Params (M)  \\
                \midrule
                GCN & $\sim$7.81M     \\
                AliNet & $\sim$16.18M\\
                IMEA  &  $\sim$20.44M   \\
                GAEA (ours) &  $\sim$8.10M  \\
                \midrule
            \end{tabular}
        \end{table}
    \end{minipage}
    \begin{minipage}[t]{0.58\textwidth}
        \begin{table}[H]
            \centering
            \scriptsize
            \caption{Ablation study results.}
            \label{tab:ablation-study}
            \begin{tabular}{lcccccc}
                \midrule
                \multirow{2}{*}{Models} & \multicolumn{3}{c}{EN-DE-15K}  &  \multicolumn{3}{c}{D-W-15K} \\
                & Hit@1 & Hit@5 & MRR & Hit@1 & Hit@5 & MRR \\
                \midrule
                GAEA & \textbf{.684}  &  \textbf{.854} & \textbf{.760} & \textbf{.562} & \textbf{.768} & \textbf{.654} \\
                $-gaal.$ & .674  &  .848 & .751 & .557 & .764 & .650 \\
                $-\mathcal{L}_c$ & .665 & .841 & .744 & .544 & .755 & .639 \\
                \midrule
            \end{tabular}
        \end{table}
    \end{minipage}
\end{minipage}

\paragraph{\rm \textbf{Ablation study.}} In the above experiments, the overall effectiveness of GAEA is proved. In this section, we conduct ablation analyses to demonstrate the validity of each component of GAEA. First, Table \ref{tab:performance-comparison} also gives the results of a variant of GAEA (denoted as w/o $rel.$), which means the original GAEA eliminates relation injection. The ablation results clearly show the effectiveness of relation embedding learning, which identifies the relation semantics can help in enriching the expressiveness of entity representations. Next, Table \ref{tab:ablation-study} gives the ablation results about graph augmentation. $-gaal.$ and $-\mathcal{L}_c$ represent the variants by removing graph augmentation in alignment learning (i.e., Eq.(\ref{eq:align-loss})) or removing contrastive objective (i.e., Eq.(\ref{eq:contrastive-loss1})), respectively (the results of removing graph augmentation are illustrated in the next section). The results show that utilizing graph augmentation can have positive impacts on EA and consistently get better performance. By introducing graph augmentation into EA training process, the model not only is encouraged to learn useful and robust entity representations but also lets the scarce yet valuable alignment seeds and vast unlabeled entities in KGs jointly provide abundant supervision for model learning.

\begin{figure}[b]
    \centering
    \includegraphics[width=\textwidth]{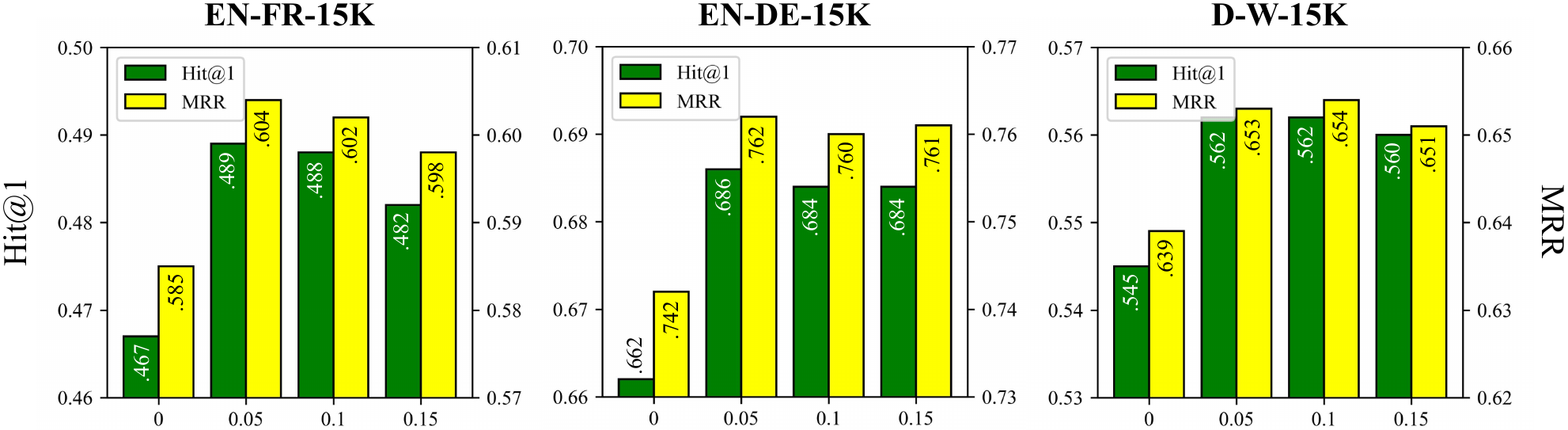}
    \caption{Parameter analysis results of $pr$ measured by Hit@1 (green bar with left axis) and MRR (yellow bar with right axis).}
    \label{fig:params}
\end{figure}

\paragraph{\rm \textbf{Parameter analysis.}} Considering that our model employs edge dropping to generate augmented graphs for margin-based alignment learning and contrastive entity representation learning. We investigate how the alignment performance varies with the upper bound of the deletion ratio. We evaluate upper bound $pr$ in \{0, 0.05, 0.1, 0.15\}, and the results measured by \textit{Hit@1} and \textit{MRR} are drawn in Fig. \ref{fig:params}. The performance is worst on all three tasks when $pr$=0, i.e., without any graph augmentation enhancement, indicating that graph augmentation can do benefit for alignment learning. We can see that the alignment effect is best when $pr$ equals 0.05 or 0.1, increasing $pr$ to 0.15 will not further improve the performance, and even bring performance drops. One potential reason is that when $pr$ becomes large, edge dropping will lead to losing more semantic knowledge and structural information, thus bringing an adverse impact on neighborhood aggregation and model training. Therefore, we need to set $pr$ as a suitably small value to ensure information retention as well as performance improvement.

\section{Discussion and Conclusion}
In this paper, we propose GAEA, a novel entity alignment method based on graph augmentation. Specifically, we design an Entity-Relation (ER) Encoder to generate latent representations for entities via jointly capturing neighborhood structures and relation semantics. Meanwhile, we apply graph augmentation to create two graph views for margin-based alignment learning and contrastive entity representation learning, thus improving the model's alignment performance. Finally, experimental results verified the effectiveness of our method.

Although GAEA achieves promising results, it still has limitations that need further investigation. First, our experimental results show that graph augmentation learning can bring some performance gains, but the supervision signals provide key performance bases in the alignment learning process. Thus, it is worth further studying how to amplify the improvement brought by graph augmentation when there no alignment seeds are given. Besides, we currently apply edge dropping as the only graph augmentation strategy, which exposes a new problem, that is, how to conduct graph augmentation learning in a highly structured KG to improve performance without introducing logic errors.

\paragraph{\rm \textbf{Acknowledgments.}} We thank reviewers for their helpful feedback. This work is supported by the National Natural Science Foundation of China No. 62172428.

\bibliographystyle{splncs04}
\bibliography{reference}

\end{document}